\documentclass[11pt,a4paper]{article}
\usepackage[hyperref]{acl2020}
\usepackage{amsmath}
\usepackage{times}
\usepackage{latexsym}
\usepackage{graphicx}
\usepackage{tabularx}
\usepackage{enumitem}
\usepackage{svg}

\usepackage{microtype}

\usepackage{tikz}
\usetikzlibrary{calc}

\aclfinalcopy

\title{Efficient Deployment of \\Conversational Natural Language Interfaces over Databases}

\author{Anthony Colas$^*$, Trung Bui$^\dagger$, Franck Dernoncourt$^\dagger$,  Moumita Sinha$^\dagger$, Doo Soon Kim$^\dagger$\\
  University of Florida$^*$, Adobe Research$^\dagger$\\
  \texttt{acolas1@ufl.edu}$^*$,\\
  \texttt{\{bui, franck.dernoncourt, mousinha, dkim\}@adobe.com}$^\dagger$\\
}

\date{}

\begin{document}
\maketitle

\begin{tikzpicture}[remember picture, overlay]
\node at ($(current page.north) + (-0in,-0.5in)$) {Published at ACL NLI 2020};
\end{tikzpicture}

\begin{abstract}
Many users communicate with chatbots and AI assistants in order to help them with various tasks. A key component of the assistant is the ability to understand and answer a user’s natural language questions for question-answering (QA). Because data can be usually stored in a structured manner, an essential step involves turning a natural language question into its corresponding query language.  However, in order to train most natural language-to-query-language state-of-the-art models, a large amount of training data is needed first. In most domains, this data is not available and collecting such datasets for various domains can be tedious and time-consuming. In this work, we propose a novel method for accelerating the training dataset collection for developing the natural language-to-query-language machine learning models. Our system allows one to generate conversational multi-term data, where multiple turns define a dialogue session, enabling one to better utilize chatbot interfaces. We train two current state-of-the-art NL-to-QL models, on both an SQL and SPARQL-based datasets in order to showcase the adaptability and efficacy of our created data.
\end{abstract}

\section{Introduction}
\label{sec:introduction}
Chatbots and AI task assistants are widely used today to help users with their everyday needs. One use for these assistants is asking them questions on various areas of knowledge or how to accomplish different tasks~\citep{braun2017evaluating, cui2017superagent}. Because data is usually stored in a structured database, in order to answer a user's questions, it is essential that the system should first understand the question, and convert it into a structured language query, such as SQL or SPARQL, to fetch the correct answer. 
\begin{figure}
  \includegraphics[width=\linewidth]{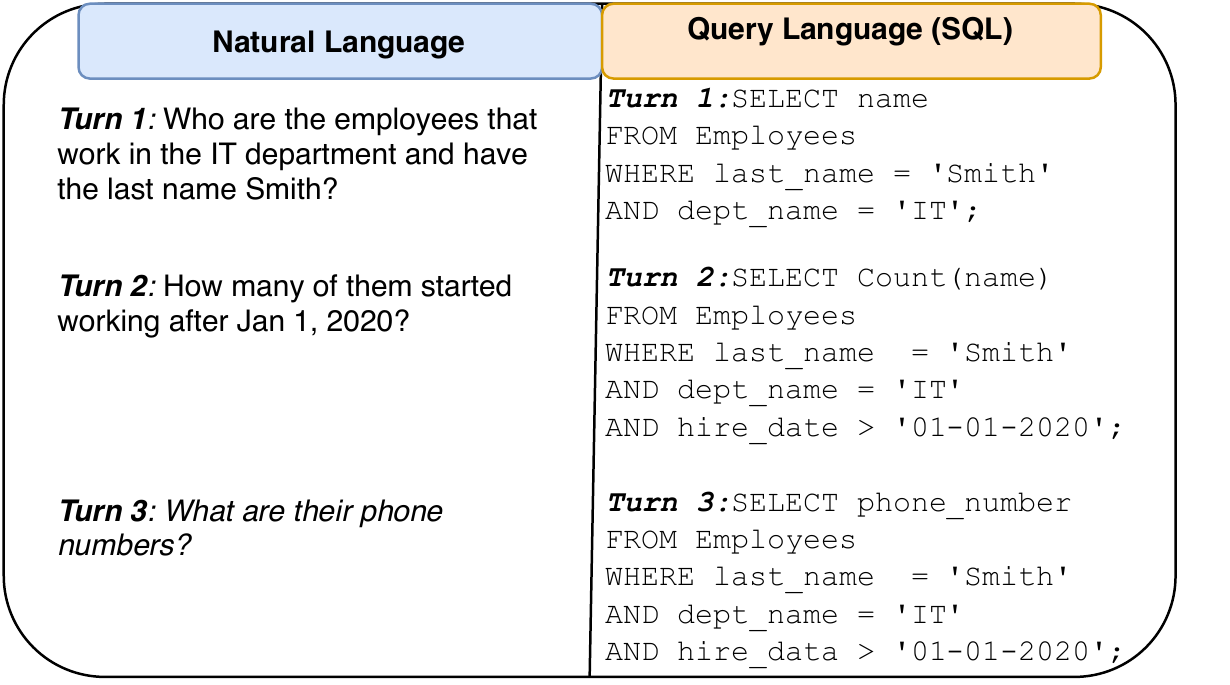}
  \caption{Example illustrating a three-turn dialogue, featuring the natural language (first column) and query language (second column) representations.}
  \label{fig:example}
\end{figure}

While much research has focused on translating natural languages into query languages~\citep{ngonga2013sorry, braun2017evaluating, dubey2016asknow, giordani2009semantic, finegan2018improving, giordani2008mapping, xu2017sqlnet, zhong2017seq2sql}, the state-of-the-art systems typically involve a large amount of training data. Therefore, in order to fully utilize these models that translate a natural language (NL) question into query language (QL), one would need to collect large amounts of both NL-QL pairs. Although there are works which involve the collection of NL-QL pairs in different domains~\citep{hemphill1990atis, zelle1996learning, zhong2017seq2sql, yu2018spider, yu2019sparc}, data is still not available in most domains, and thus this collection process can be both time-consuming and expensive. 

In this work, we address the problem of having insufficient data collection methodologies by proposing a novel approach that accelerates the data collection process for use in NL-to-QL models. Additionally, our approach focuses on generating conversation data, where the context of a dialogue turn is used to generate a subsequent pair. In this way, we better simulate the data necessary for real world chatbots and voice assistants, as exemplified in Figure~\ref{fig:example}. Our contributions are as follows: 
\begin{itemize}
  \item We develop a novel approach that accelerates the creation  of NL-to-QL data pairs. Primarily, our approach tackles the problem in the conversational domain. 
  \item We showcase our data collection system on two different QLs, SQL and SPARQL, demonstrating the flexibility of our system.
  \item Finally, we demonstrate the use of current single-turn state-of-the-art approaches on these two domains to prove the adaptability of our system to current models.
\end{itemize}

Though our data collection implementation focuses on conversational data, the models we deploy are single-turn. Our main focus here is to give a demonstration of the generated data. Section 3 and Section 4 show the adaptability of our data collection scheme to these kinds of models. 

The rest of this paper is structured as follows: Section~\ref{sec:related} surveys prior work in both the NL-to-QL and data collection space, Section~\ref{sec:data_collection} details our novel conversational data collection approach, Section~\ref{sec:data} walks through examples in both the SQL and SPARQL domain, Section~\ref{sec:experiments} describes the current models we have trained and tested on the generated data, Section~\ref{sec:results} gives the results on the data and models, and Section~\ref{sec:conclusion} concludes our work. 

\section{Related Work}
\label{sec:related}
In the field of natural language interfaces for structured data there are bodies of work that 1) focus on translating natural language to a specific query language and that 2) relate to collecting semantic parsing data for natural language interfaces. 

\subsection{NL-to-QL}
NL-to-QL models have worked to transform natural language queries into their respective logical form (LF) representations~\citep{dong2016language}, SQL queries~\citep{xu2017sqlnet,zhong2017seq2sql,finegan2018improving,cai2018encoder}, or SPARQL queries~\citep{ngonga2013sorry,dubey2016asknow}. While work in the SPARQL domain first normalize and match the queries, state-of-the-art work in translating NL to SQL involves neural architectures. \citet{dong2016language} utilize and encoder-decoder framework to translate NL questions into their LF representation. \citet{xu2017sqlnet} propose a sketch-based model where a neural network predicts each slot of the sketch. The architecture built by \citet{zhong2017seq2sql} uses policy-based reinforcement learning in order to translate NL to SQL. While \citet{finegan2018improving}'s main takeaway is how different evaluations effect the generalization problem in translating NL to SQL, they approach the problem with a seq2seq model. Because of the volume of data needed to fully utilize these models, it can be difficult to adapt to different domains.

In the multi-turn domain, \citet{saha2018complex} first approach the problem of complex sequential question-answering (CSQA) by first building a large-scale QA dataset made to answer questions found in Wikidata~\footnote{https://www.wikidata.org/wiki/Wikidata:Main\_Page}. However, their data collection process was extremely laborious, as their process required in-house annotators, crowdsourced workers, and multiple iterations. Additionally, their approach was end-to-end, meaning the output was an expected answer. Nevertheless, because their approach incorporate the query representation, we plan to further incorporate their approach into our data collection process in future work.\citet{yu2019cosql} also develop the first general-purpose DB querying dialogue system. However, their system dialogues focus on clarifying a NL question for user verification, before returning an answer. Our work focuses on generating conversational data about specific database entities and properties. 

\subsection{Data Collection for Semantic Parsing}
NL question semantic parsers have been developed for single-turn QA in order to translate simple NL questions into their respective LFs~\citep{wang2015building}. In their approach, \citet{wang2015building} first begin with a \textit{domain}, building a seed lexicon of that domain. Next, they find the LF and canonical utterance templates corresponding based on the lexicon. \citet{wang2015building} then paraphrase their canonical utterances via crowd-sourcing. \citet{iyer2017learning} learn a semantic parser via an encoder-decoder model by using NL/SQL templates. This model is tuned through user feedback, where incorrect queries are annotated by crowd-workers. Paraphrasing is accomplished through the Paraphrasing Database (PPDB)~\citep{ganitkevitch2013ppdb}.

While the two previously mentioned works are single-turn semantic parsers, \citet{shah2018building} develop a multi-turn semantic parser. Their approach begins with a task schema and API which is used to create dialogue outlines for the provided domain. These dialogue outlines involve a user and system bot that simulate a scenario. The dialogues are then paraphrased via crowd-sourcing. However, \citet{shah2018building} use the logical-form representation of the utterances rather than their query language representation. In our work, we re-incorporate the paraphrases into the dialogue generation phase. 

\section{Data Collection System}
\label{sec:data_collection}
Our conversational data collection strategy is developed to efficiently collect NL/QL pairs for training data in models which translate the NL into QL in a multi-turn setting. Because domain data is required when training a chatbot to query a database when converting from NL to QL, our approach is generalized so that one can easily collect data for their respective domain. 

\subsection{Overview}
\begin{figure}
  \includegraphics[scale=1.0]{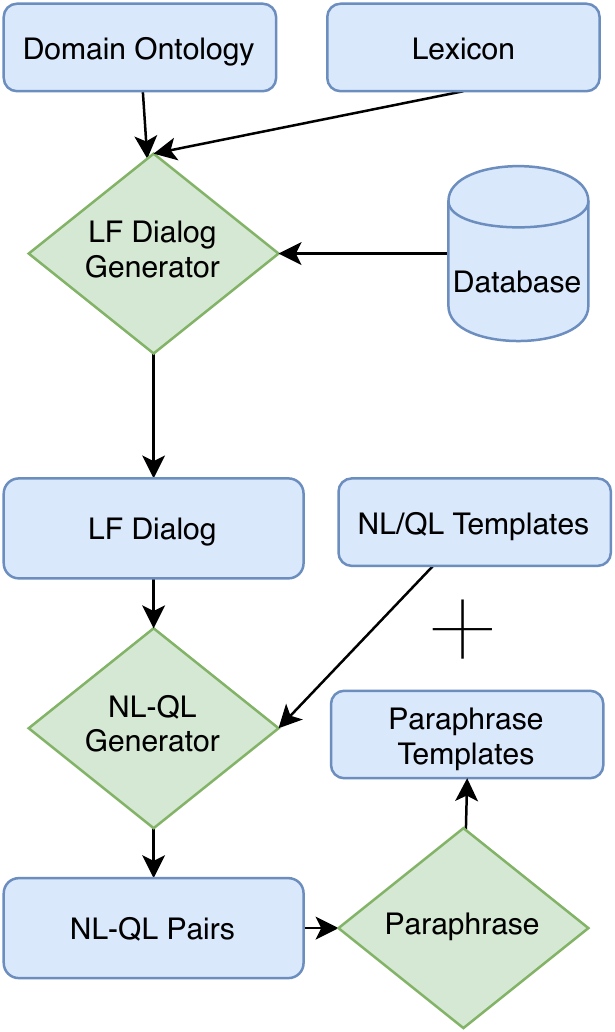}
  \caption{An overview of our conversational data collection deployment system. Blue shapes denote the input/output data at each stage, while green diamonds denote the processes of the system. The ``plus'' sign denotes the concatenation of both seed templates and paraphrase templates.}
  \label{fig:data_collection}
\end{figure}
Our approach in collecting data is made of the four following steps: 1) First we generate the dialogue represented as LFs, forming the abstract representations of NL questions, 2) Next, we convert the LFs into an NL template and QL templates 3) We then collect paraphrases of the natural language templates, and 4) Finally, we use these paraphrases to further develop our dialog generator. In generating our dialogue, the context of each previous turn is taken in order to develop the current turn. Figure~\ref{fig:data_collection} presents our data deployment system. We divide and expand upon the steps further in the next sections.

\subsection{Definitions}
We first define the following notations in our data collection system:
\begin{itemize}
\item $U_n$: an utterance in the dialogue.
\item $LF_n$: the LF \textit{n} in the dialogue.
\item $NL_n$: the NL utterance corresponding to $LF_n$.
\item $QL_n$: the QL utterance corresponding to $LF_n$.
\end{itemize}

\subsection{Input Module}
The input to our data collection system consists of a domain ontology, lexicon, and database. These should be provided by the user and vary depending on the type of data one requires. The domain ontology defines the \textit{$<$object, relation, property$>$} triples of a given dataset, where each object has a set of properties connected through a relation, e.g. \textit{$<$ACL 2020, has\_location, Seattle$>$}. The lexicon file defines each data field, along with its NL and QL representation, important in the NL-QL Generator step. The database is the data in structured form.

\subsection{Logical Form Dialogue Generator}
In order to appropriately simulate a conversation between a user and chatbot, the synthetic dialogue must first be generated. This is done by first outlining the dialog via LFs, where the system generates, $LF_{1-n}$. These outlines are an abstract but understandable representation of the dialogue taking into account the type, entity, and relation of a question. Thus, our parser builds a dialogue based on a domain ontology, lexicon, and domain database.

The LFs take the form of three predicates: \textit{Retrieve-Objects}, \textit{Inquire-Property}, and \textit{Compute}, each taking on their own arguments. For the  \textit{Retrieve-Objects} predicate, the LF fetches an instance that satisfies a condition. As arguments, \textit{Retrieve-Objects} takes an \textit{entity type}, $\mathbf{t^i_n}$ from the ontology, a boolean \textit{condition} $\mathbf{c^i_n}$, and a \textit{property value},$\mathbf{p^i_n}$, from the DB. For the \textit{Inquire-Property} predicate, given an \textit{anchor entity} $\mathbf{ae^i_n}$, \textit{target instance}, $\mathbf{ti^i_n}$, and an \textit{inference path} $\mathbf{ip^i_n}$ from the entity to that instance, the LF finds the property in that path of the anchor entity. The \textit{Compute} predicate denotes a \textit{computation} $\mathbf{comp^i_n}$ over a set of given objects, thus its arguments are comprised of \textit{Retrieve-Objects} arguments and an operation to be performed.~\footnote{\textbf{n} refers to the dialogue turn, while \textbf{i} refers to the number of dialogue generated.}. For our work, we focus on using the \textit{COUNT} aggregate function. Future work can easily adapt more aggregate functions into our model such as \textit{MAX} or \textit{MIN} depending on the values contained in the database. 

More formally, each LF can be described as follows:
\begin{subequations}\label{eqn}
\begin{align*}
\boldsymbol{LF_n} \rightarrow \{Retrieve-Objects(t^i_n,c^i_n,p^i_n),\\ Inquire-Property(ae^i_n,ti^i_n,ip^i_n), \\
Compute(comp^i_n, t^i_n,c^i_n,p^i_n)\} \tag{\ref{eqn}} 
\end{align*}
\end{subequations}

At the start of a dialogue, a random LF predicate is selected, given the database schema, lexicon, and domain ontology. The subsequent turns in the dialogue are built conditionally on the previous turn. Therefore, given a $LF_{n-1}$, when generating $LF_n$ the context of $LF_{n-1}$ is further taken into consideration including its arguments, type, and answer. The subsequent predicate is also chosen at random, however its values are conditional on the arguments and answer(s) of the current predicate. For example, if $LF_{n-1}$ is an \textit{Retrieve-Objects} predicate and another \textit{Retrieve-Objects} predicate is chosen as $LF_n$, this LF can further filter the answer of $LF_{n-1}$ by using an additional condition. 
\begin{table*}[t]
\centering
\small
\begin{tabular}{|l|l|l|l|}
\hline
\textbf{Predicate} & \textbf{Explanation}           & \textbf{Example}                                                                            & \textbf{LF}                                                                                                                      \\ \hline
Retrieve-Objects   & Gets objects from DB           & \begin{tabular}[c]{@{}l@{}}Which employees have \\ building no. equal to 5?\end{tabular}    & \begin{tabular}[c]{@{}l@{}}Retrieve-Objects(employee(ALL), \\                            (employee.building\_no,`=',5))\end{tabular} \\ \hline
Inquire-Property   & Gets an object's property      & What is the office of James?                                                                & Inquire-Property(James, office)                                                                                                  \\ \hline
Compute            & CompuAggregate function & \begin{tabular}[c]{@{}l@{}}How many employees have \\ hire year equal to 2010?\end{tabular} & \begin{tabular}[c]{@{}l@{}}Compute(COUNT, employee(ALL), \\                {[}('hire\_year', 2010){]})\end{tabular}              \\ \hline
\end{tabular}
\caption{LF predicate summary with an explanation and example of each, both in NL and LF.}
\label{tab:LF Predicate Summary}
\end{table*}
Table 1 summarizes the types of LFs, along with an explanation and example of each both in LF and NL, which we discuss in the next section.

\subsection{NL-QL Generator}
\label{sec:nl-ql_generator}
Once the LF generator is complete, the data collection system generates an NL utterance along with its corresponding QL. To generate such pairs, the NL-QL generator takes in each LF from the LF Dialog as input. Based on the predicate type, an NL-QL pair is selected and filled with corresponding arguments of the predicate. Thus, the system uses NL seed templates for the \textit{Retrieve-Objects}, \textit{Inquire-Property}, and \textit{Compute} predicates to create the initial training data for the conversational dialogue. For example, one NL template for turns after \textit{$NL_1$} can be \textit{"How about \textless{}entity\textgreater?"}

The aforementioned seed templates are hand-crafted based on the type of data and are thus left to the user to create. These data are hand-crafted to increase the quality of the seed templates in terms of coherency and utility, important features not only for quality training data, but also when performing the paraphrase task. Because we hand-crafted the query language templates, we also guarantee that the queries are executable for their corresponding QLs, SQL or SPARQL in this work. For the QL, we fill in slots for field names, aliases, and values, utilizing the information in the domain ontology, lexicon, and database schema. Note, `field' refers to column names in relational DBs (queried with SQL) and type names in graph DBs (queried with SPARQL). To reiterate, the NL-QL generator takes each $LF_n$, with its respective arguments, and seed templates as input, and outputs a $NL_n-QL_n$ pair, where $U_n \rightarrow (NL_n,QL_n)$.
Section~\ref{sec:data} goes through detailed examples of various NL-QL pairs.

\subsection{Paraphrase}
The final step involves the paraphrasing of the seed NL templates given in the NL-QL Generator step. To paraphrase the seed NL templates, we first provide crowdworkers from Amazon Mechanical Turk (AMT)~\footnote{https://www.mturk.com/} with the instantiated templates, the output from the first iteration of the NL-QL generator. We ask the workers to paraphrase the seed templates while keeping the meaning/intent of the original questions. After collecting these paraphrased questions, we further abstract them and link them to their respective predicate representation. In this way, the paraphrases can be utilized in further iterations of the NL-QL Generator step and instantiated when generating new dialogues for training data. While abstracting the templates, we manually scan them for quality control purposes. Furthermore, we ran multiple trial runs in presenting the problem to the AMT workers. Previous work~\citep{wang2015building,shah2018building} also use similar crowd-sourcing techniques in order to paraphrase their templates. Via AMT, \citet{wang2015building} paraphrase canonical utterance, natural language representations to single-turn LFs, while \citet{shah2018building} paraphrase dialogue outlines as their final step.

Similarly to \citet{shah2018building}, we input the paraphrases back into our NL-QL generation step. Figure~\ref{fig:data_collection} illustrates this through the ``+" symbol, signifying that the paraphrases are appended to the seed templates when mapping to LF and creating the final NL-QL pairs. This approach can take multiple iterations, as the user sees fit to the NL question generation task in their data domain. 

\section{Data Examples}
\label{sec:data}
In this section we will showcase examples in both the SQL and SPARQL domain and traverse through each stage of our Data Collection System. We first begin with SQL, used to query relational databases, and then demonstrate our system with a graph querying language, SPARQL. By doing so, we show the extendability of our approach to various structured QLs. Moreover, we confirm the importance of generating executable queries in a conversational data collection system.

\subsection{SQL}
\label{sec:SQL}
\begin{figure}
  \includegraphics[width=\linewidth]{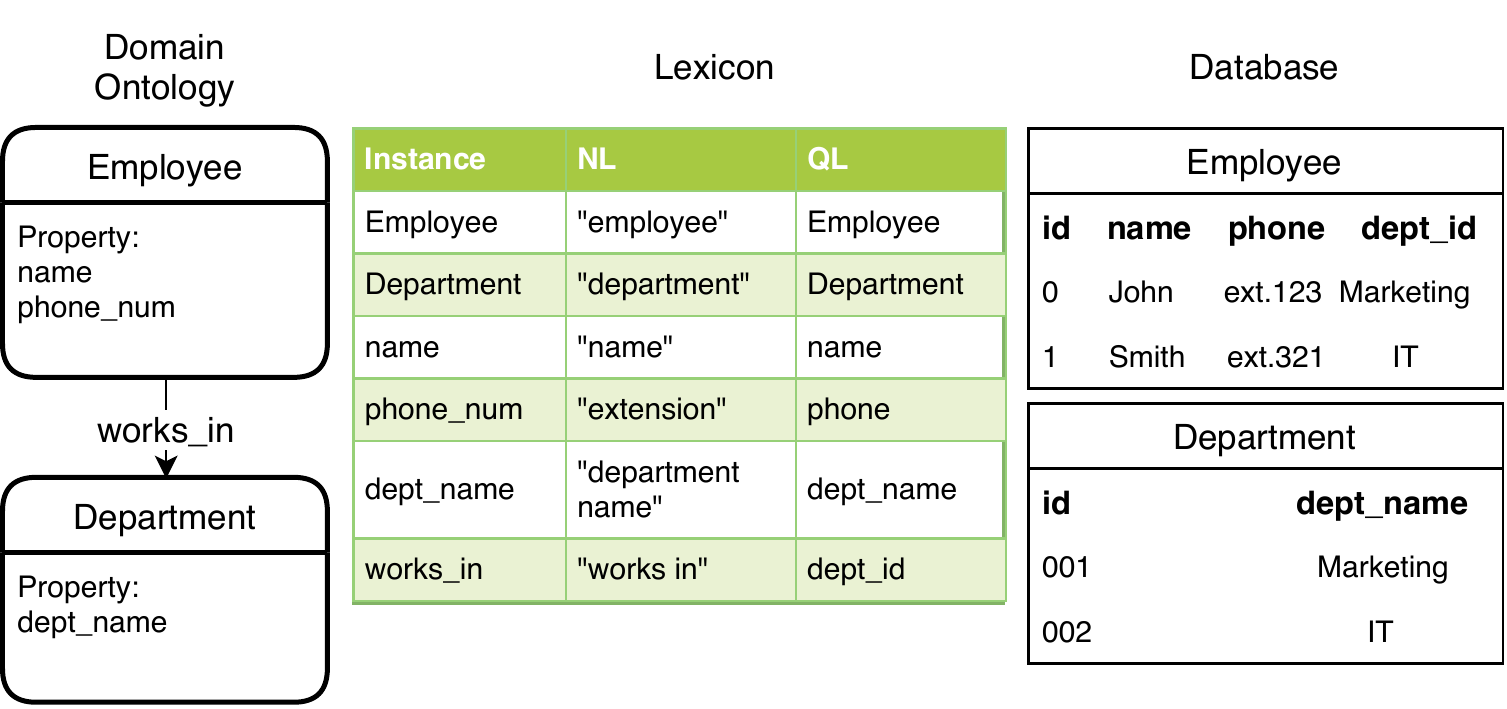}
  \caption{Example ontology schema, lexicon, and database. The two tables in the Database are used throughout our SQL example.}
  \label{fig:input}
\end{figure}

Through our data collection system for conversational QA, we are able to produce contextual dependent NL-SQL pairs. For the SQL example, suppose a user wants to produce data for an employee directory relational database. Figure~\ref{fig:input} gives an example of possible input files needed to produce this kind of conversational data with our data collection system, including a domain ontology with two entities \textit{Employee} and \textit{Department}, a lexicon to map NL and QL instances, and a database containing \textit{Employee} and \textit{Department} data. 

Thus, given the input files in Figure 3, possible $LF_n$ values with each predicate are:
\begin{enumerate}[label=(\roman*)]
\item Retrieve-Object(employee(ALL),\\
    (employee.dept\_name,`=', Marketing))
\item Inquire-Property(James,dept\_name)
\item Computation(COUNT,employee(ALL),
    [(`works\_in', `IT')])
\end{enumerate}
In (i), the logical form represents a retrieval of employee objects who work in the Marketing department. (ii) asks about the department name of James. (iii) computes the total number of employees who work in the IT department. During the generation of LF\_1, one of these LFs can be generated. Then for LF\_2 - LF\_n, the context is passed along to generate the LFs. The \textit{n} denotes the number of turns a dialogue can take. As an example, given LF\_1 is (1) from the aforementioned LFs, LF\_2 can be \textit{Inquire-Property(Answer,phone\_num)}, where \textit{Answer} denotes the objects returned by LF\_1. Our dialogue generation system allows one to tune the number of turns and number of dialogues generated from the given input. 

For the NL-QL step, our input includes the dialogues represented as LFs along with NL-QL seed templates described in Section~\ref{sec:nl-ql_generator}. Possible templates are given in Table~\ref{tab:templates}. Note, that we refer to a column in a relational DB as a field. Taking our previous \textit{Retrieve-Objects} example, the filled seed template would read: ``Which employee have department equal to Marketing?" The Lexicon from Figure~\ref{fig:input} is utilized here, as the instance name is mapped to its NL name. Similarly, its QL name (table name) is mapped in the SQL query.

\begin{table}[t]
\small
\begin{tabular}{|l|l|}
\hline
\textbf{Predicate} & \textbf{Template}                                                                                                                                                       \\ \hline
Retrieve-Object    & \begin{tabular}[c]{@{}l@{}}Which \textless{}entity\textgreater have \textless{}field name\textgreater \\ equal to \textless{}instance\textgreater{}?\end{tabular}   \\ \hline
Inquire-Property   & \begin{tabular}[c]{@{}l@{}}What is the \textless{}field name\textgreater\\  of \textless{}entity value\textgreater{}?\end{tabular}                                      \\ \hline
Computation        & \begin{tabular}[c]{@{}l@{}}How many \textless{}entity\textgreater  have\\ \textless{}field name\textgreater equal to \textless{}instance\textgreater{}?\end{tabular} \\ \hline
\end{tabular}
\caption{Examples of seed templates with their respective predicates. \textless{}entity\textgreater refers to an entity type. \textless{}field name\textgreater corresponds to a column in a relational DB or a relation in a graph DB. \textless{}instance\textgreater refers to the value of that field in the DB. \textless{}entity value\textgreater is an instance of an entity in the DB.}
\label{tab:templates}
\end{table}

Finally, in the final step, as explained in~\ref{sec:nl-ql_generator}, the NL seed templates are paraphrased via crowd-sourcing, e.g. ``Which employee have department equal to Marketing?" can be paraphrased into ``Who works in the marketing department?". 

\subsection{SPARQL}
\label{sec:sparql}
\begin{figure}
  \includegraphics[width=\linewidth]{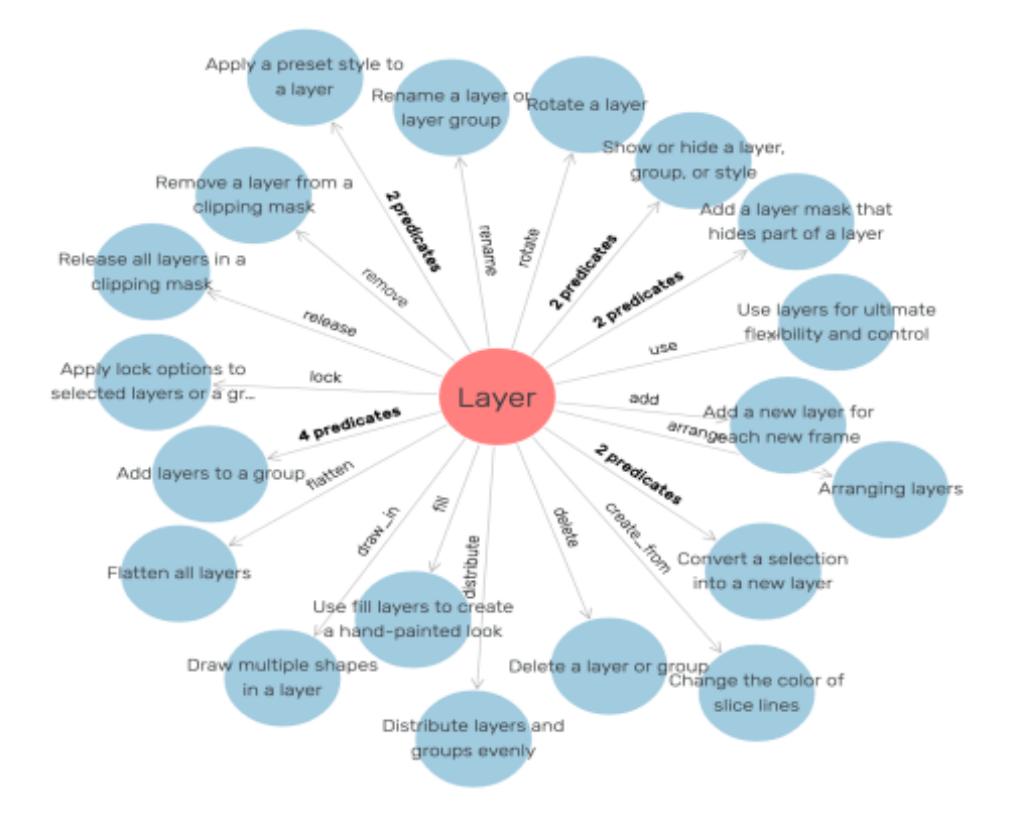}
  \caption{An example of a subgraph in the Photoshop Knowledge Graph. The Layer object (red node), can be seen connected to its objects (blue nodes) through relations. Here we can see that the Layer entity is connected to the various actions connected to ``Photoshop Layers", such as ``flatten", ``lock", and ``use", where the object nodes show how they can be performed.}
  \label{fig:photoshop}
\end{figure}

SPARQL is used to query graph databases, where entities are linked together through relations. These graph databases usually take the form of triples in the form: \textit{\textless{subject,relation,object}\textgreater}. Because both LF-Generator and NL-QL Generator remain the same as in Section~\ref{sec:SQL}, here we examine the main differences in the system data when utilizing SPARQL instead of SQL. As a guide, we refer to the example give in Figure~\ref{fig:photoshop}.

Figure~\ref{fig:photoshop} gives an example of a subgraph found in the Photoshop Knowledge Graph (KG). This KG contains the various tools, dialogs, shortcuts, and options found in Photoshop, connected to their options and definitions through relations. The KG is extracted from the Photoshop Wiki. Similarly to the SQL example above, we input a domain ontology, lexicon, and database to the conversational data collection system. However, in the case of a graph database, the entities found in the ontology are more clearly defined in a graph database. Additionally, instead of a table structure, the database is in the form of \textit{\textless{subject,relation,object}\textgreater} triples, where each entity belongs to a \textit{type} defined in the ontology.

While the the types of LFs generated in the LF-Generator are equivalent, a \textit{property} now refers to the relation found in the triple, while a property refers to the object of a KB triple. For example, an entity such as the one found in figure~\ref{fig:photoshop} may have various properties, including ``has\_shortcut'' and ``has\_option''. When generating NL-QL pairs, the generator again takes from the out of the LF-Generator, lexicon, and seed templates, where the QL template is SPARQL-based instead of SQL-based. Paraphrases are collected in the same way. Thus, an example Photoshop Retrieve-Object LF template question, and paraphrase may look like: ``LF: Retrieve-Objects(tool(ALL),
    (tool.has\_shortcut, ‘=’, ‘H))", ``Template: Which $\langle$ entities$\rangle$ have $\langle$ relation$\rangle$ equal to $\langle$ object$\rangle$?", and ``Paraphrase: What's the tool with the H shortcut?"

\section{Experiments}
\label{sec:experiments}
We will now examine our experiments with a relational and graph database setting. We first briefly discuss the data used in constructing the converstational dataset and then describe the various models utilized in translating the NL questions into their respective structured queries. 

\subsection{Data}
For our experiments involving SQL data, we construct an NL-QL conversational dataset on data based on a proprietary web analytics tool. In our results table, we refer to this dataset as \textit{Web-Analytics}. For the graph-database, we construct an NL-QL conversational dataset based on the Photoshop KB, as the one exemplified in Section~\ref{sec:SQL}. As previously noted, this KB contains various entities found in Photoshop, connected to their properties, through predicates which define the properties. In total, the KB contains 15,381 triples, with 3,410 triples that correspond to how-to type queries. 

After running our conversational data collection system on both set of data, we collected 288 and 73 NL-QL pairs of templates for the Photoshop and Web-Analytics datasets, respectively. Table~\ref{tab:template_stats} summarizes these statistics. Additionally, we configured our system to give 3 turn dialogues.
\begin{table}[]
\begin{tabular}{|l|l|l|}
\hline
          & Photoshop & Web-Analytics \\ \hline
Templates & 288       & 73            \\ \hline
\end{tabular}
\caption{Number of templates for each dataset, where the Photoshop dataset is SPARQL-based and Web-Analytics dataset is SQL-based.}
\label{tab:template_stats}
\end{table}

\subsection{Models}
In our experiments we utilize single-turn NL-QL models. Specifically, we utilize the baselines defined by \citet{finegan2018improving}.

The first baseline is a seq2seq model with attention-based copying, originally proposed by \citet{jia2016data}. This model takes an NL utterance as input and outputs a structured query. Included in the output is a COPY token, which signifies the copying of an input token. In the copying mechanism model, the loss is calculated based on the accumulation of both the probability of distribution of the tokens in the output and the probability of copying from an input token. This copying probability is calculated as the categorical cross entropy of the distributed attention scores across the input's tokens, where the token with the max attention score is chosen as the output token. 

\begin{figure}
  \includegraphics[width=\linewidth]{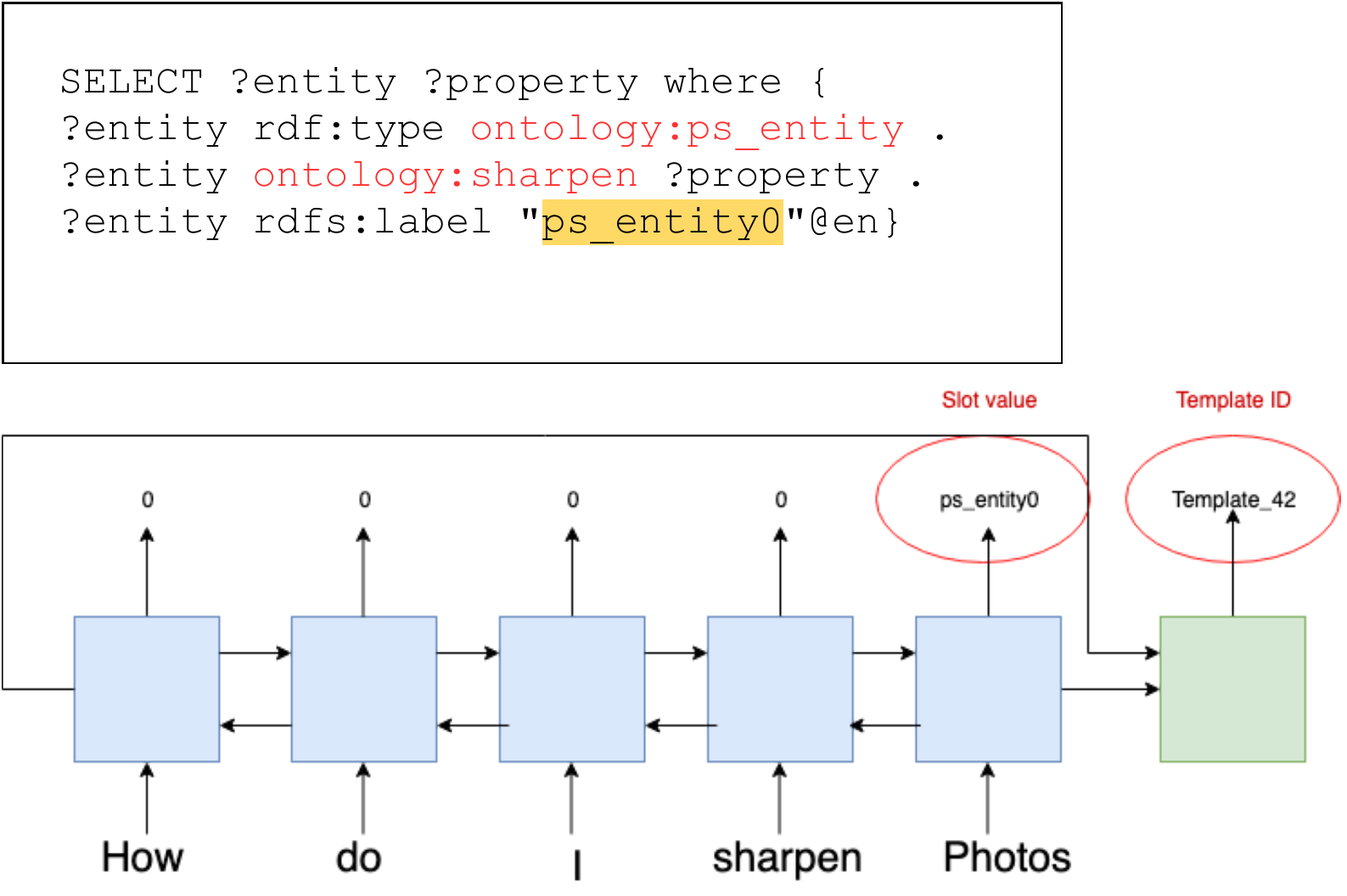}
  \caption{The template-based model developed by \citet{finegan2018improving}, where the blue boxes represent LSTM cells and the green box represents a feed-forward neural network. `Photos' is classified as a slot value, while the template chosen (Tempalte\_42), is depicted above the model. In the template, the entity slot is highlighted in yellow and the properties which make the template unique are in red.}
  \label{fig:template-based}
\end{figure}

The second baseline is a template-based model developed by \citet{finegan2018improving}. This model takes in natural language questions, along with query templates to train. Since our data collection system directly utilizes templates to generate the data, this model is easily adaptable to our setting. We simply use the templates we collect from both the seed-templates and paraphrasing tasks, as well as the slot values extracted from the source DB when creating the dialogue data to train the model. In the template-based model, there are two decisions being made. First the model selects the best template to choose from the input. This is done by passing the final hidden states of a bi-LSTM through a feed-forward neural network. Next, the model selects the words in an input NL-question which can fill the template slots. Again, the same bi-LSTM is used to predict whether an input token is used in the output query or not. Thus, given a natural language question, the model jointly learns the best template from the given input, as well as the values that fill the template's slots. Please note, that while this model is best fitted for our dataset, it does not generalize well to data outside of the trained domain due to the template selection task. Figure~\ref{fig:template-based}, inspired by \citet{finegan2018improving}, shows an example of the template-based model with our own input in the SPARQL domain. 

Although our dataset collection system generates multi-turn data, because of the immaturity of multi-turn NL-to-QL models, we leave the use of multi-turn models for future work. We do however, mention the model developed by \citet{saha2018complex}, which answers complex sequential natural language questions over KBs, which can be further integrated in future work.
\begin{figure*}
  \includegraphics[width=\linewidth]{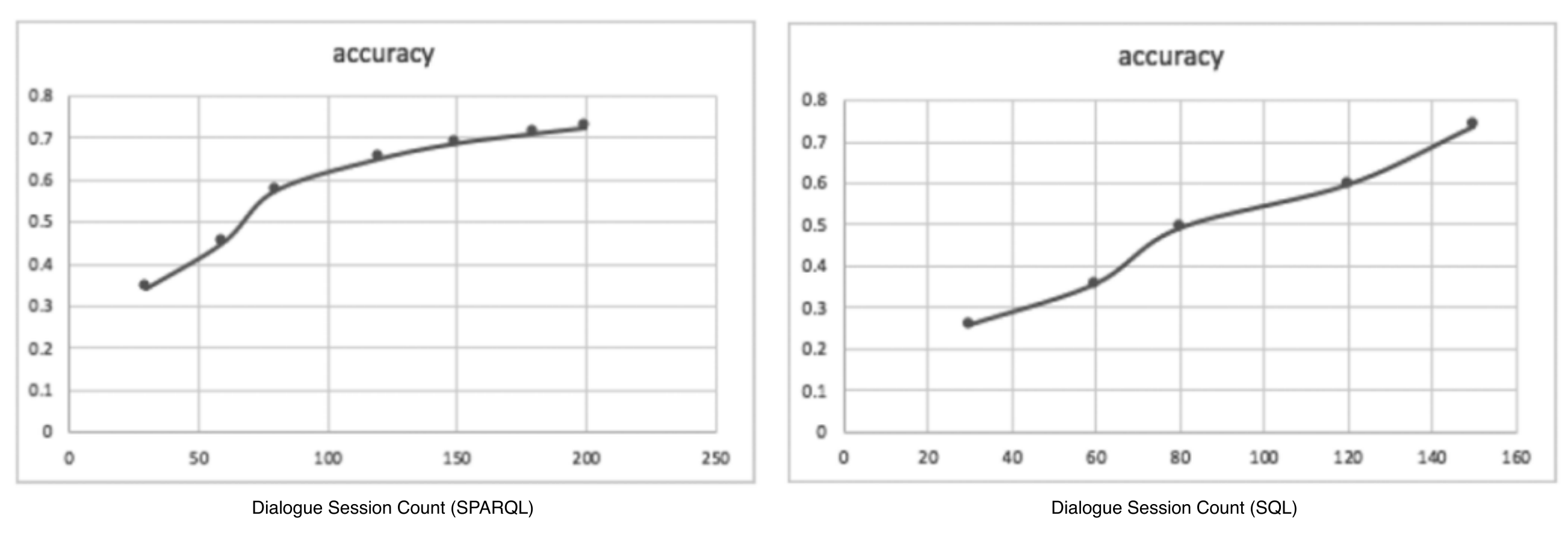}
  \caption{The above graphs show that as the dialogue session count increases for both the Photoshop SPARQL (left) and Web-Analytics SQL (right) dataset, the accuracy also increases. The y-axis of each graph marks the accuracy, while the x-axis marks the number of dialogue sessions for each dataset.}
  \label{fig:results}
\end{figure*}
\subsection{Settings}
We experimented with both the seq2seq and template-based models on the SQL-based and SPARQL-based datasets previously discussed. For the Photoshop SPARQL dataset, we generated \textit{2,100} single-turn data pairs utilizing our data collection system, while generating \textit{3,504} single-data pairs for the web-analytics dataset. Experiments all used a 90/10 train/validation set split.

\section{Results}
\label{sec:results}
\begin{table}[]
\begin{tabular}{l|l|l|}
\cline{2-3}
                                              & \textbf{Photoshop} & \textbf{Web-Analytics} \\ \hline
\multicolumn{1}{|l|}{\textbf{Seq2seq}}        & .726               & .738                   \\ \hline
\multicolumn{1}{|l|}{\textbf{Template-based}} & .305               & .641                   \\ \hline
\end{tabular}
\caption{Results on the accuracy of the NL-to-QL task on the generated single-turn Photoshop and Web-Analytics datasets.}
\label{tab:results}
\end{table}
We evaluated the models on our generated datasets for exact-match accuracy of the SQL/SPARQL output queries. The results (shown in Table~\ref{tab:results}) indicate that in both cases the seq2seq model outperforms the template-based model. While the seq2seq gives an accuracy of .726 and .738, the template-based model results in .305 and .641 accuracy. Furthermore, the template-based model performs better on the Web-Analytics SQL-based dataset. This may be because the number of templates contained in the SQL dataset is almost four times greater than the number of templates contained in the Photoshop SPARQL dataset, 73 compared to 288.

We also investigate how the accuracy of the models increase, as the number of samples generated by our data collection system increase. Figure~\ref{fig:results} shows that for our best performing model (seq2seq), as the number of dialogue sessions (or data points) increases, the accuracy increases. While this is expected, it also shows that through out dialog creation system, one can improve their NL-to-QL application's performance by configuring the data creation system with more dialogues and templates. 

Though the models use synthetic data generated by our system, our system allows one to accelerate the data collection process and quickly deploy an NL-to-QL system that gives reasonably accurate results. This deployed system can then later collect data collected from real application users, where the application logs where a correct or incorrect response may have been returned.~\citet{iyer2017learning} explore this kind of work which learns from user feedback, where users marked utterances as correct or incorrect, and the accuracy of the semantic parser increased as a result.

\section{Conclusion}
\label{sec:conclusion}
In this work, we propose a conversational data collection system which accelerates the deployment of conversational natural language interface applications which utilize structured data. We describe the three main processes of our system, including the \textit{LF Dialog Generator}, the \textit{NL-QL Generator}, and the \textit{Paraphrase} component. By taking in a domain ontology, lexicon, and structured database as input, our system generates NL-QL multi-turn pairs which can be used to train systems that translate NL to QL. Each component of our system is examined in both the SQL and SPARQL QL domain. We then validate our data by training state-of-the-art NL to QL models on single-turn utterances. Our experiments show promising results in both the SQL and SPARQL domains, while providing an efficient method to generate data for the development of multi-turn models.
\bibliography{acl2020}
\bibliographystyle{acl_natbib}
\end{document}